\documentclass{article} % For LaTeX2e
\usepackage[preprint]{colm2026_conference}

\usepackage{microtype}
\usepackage{hyperref}
\usepackage{url}
\usepackage{xspace}
\usepackage{booktabs}
\usepackage{amsmath}
\usepackage{graphicx}
\usepackage[table]{xcolor}
\usepackage{subcaption}
\usepackage{multirow}
\usepackage{algorithm}
\usepackage{algpseudocode}
\usepackage{etoolbox}
\AtBeginEnvironment{algorithmic}{\setlength{\itemsep}{0.5pt}}

\usepackage{lineno}

\definecolor{darkblue}{rgb}{0, 0, 0.5}
\hypersetup{colorlinks=true, citecolor=darkblue, linkcolor=darkblue, urlcolor=darkblue}

\title{TRIMS: Trajectory-Ranked Instruction Masked Supervision for Diffusion Language Models
}

\author{
Lingjie Chen\thanks{Correspondence to \texttt{lingjie7@illinois.edu}.} \\
UIUC
\And
Ruizhong Qiu \\
UIUC
\And
Yuyu Fan \\
Independent Researcher
\And
Yanjun Zhao \\
UIUC
\And
Hanghang Tong \\
UIUC
}

\newcommand{\methodname}{TRIMS}  % e.g. Rewarded SFT, Ours, ...
\newcommand{\method}{\textbf{\texttt{\methodname}}\xspace}
\newcommand{\fullmethod}{\textbf{T}rajectory-\textbf{R}anked \textbf{I}nstruction \textbf{M}asked \textbf{S}upervision\xspace}

\begin{document}

\ifcolmsubmission
\linenumbers
\fi

\maketitle

\begin{abstract}
Diffusion language models (DLMs) offer a promising path toward low-latency generation through parallel decoding, but their practical efficiency depends heavily on the decoding trajectory. In practice, this advantage often fails to fully materialize because standard training does not provide explicit supervision over token reveal order, creating a train--inference mismatch that leads to suboptimal decoding behavior. We propose \fullmethod{} (\method{}), a simple trajectory-guided supervised fine-tuning framework that injects trajectory supervision into standard Masked Diffusion Language Model (MDLM) training with minimal overhead. Instead of relying on costly DLM-based distillation, \method{} uses lightweight signals from an autoregressive teacher to guide a trajectory-aware masking strategy, encouraging the model to learn more effective decoding orders. Experiments on LLaDA and Dream across math and coding benchmarks show that \method{} significantly improves the accuracy-parallelism trade-off over both standard MDLM training and train-free acceleration baselines, while achieving competitive performance with prior distillation-based approaches at substantially lower training cost. Further analysis shows that \method{} leads to better decoding trajectories, validating the effectiveness of trajectory-guided supervision for DLMs.
\end{abstract}

\section{Introduction}

Diffusion language models (DLMs) offer a promising alternative to autoregressive (AR) language models~\citep{austin2021structured,lou2023discrete,sahoo2024simple,shi2024simplified,arriolablock}. Unlike AR models, which generate tokens in a fixed left-to-right order, DLMs can generate tokens in arbitrary orders, enabling global iterative refinement~\citep{wang2025remasking,havasi2025edit}. Moreover, the diffusion paradigm supports parallel token generation, which is a key advantage over AR models and has motivated a growing line of efficient decoding methods~\citep{wu2025fastdllm,wu2025fastdllmv2,ma2025dkv,ben2025accelerated}. Recent work on scaled DLMs has demonstrated their potential to achieve higher throughput and lower latency than AR models~\citep{labs2025mercuryultrafastlanguagemodels}.

However, in practice, the benefits of parallel decoding do not always fully materialize. Models may fall back to next-token prediction or deliberately skip difficult tokens during inference~\citep{ni2026flexibility, fu2025bits}, substantially limiting parallelism and wasting the parallel capacity afforded by bidirectional attention. Ideally, the model should learn to organize its decoding trajectory to maximize both parallelism and efficiency.

A key reason behind this limitation is the \emph{mismatch between training and inference}. Existing DLMs are typically trained with a uniform random masking strategy~\citep{sahoo2024simple,shi2024simplified}. Under this training paradigm, the noise schedule is sample-agnostic, each token is masked with the same probability, and the model is trained by optimizing the evidence lower bound (ELBO). As a result, the model does not receive explicit supervision about which tokens should be resolved earlier and which can be left to later steps. Recent works have begun exploring trajectory-aware training for DLMs, but many of them rely on costly trajectory sampling from the DLM itself for distillation~\citep{chen2025dparallel,qian2026d3llm}. While effective, such methods inherit the high cost of diffusion sampling, limiting their practicality.

Motivated by this mismatch and the limitations of existing methods, we propose \fullmethod{} (\method{}), a simple trajectory-guided supervised fine-tuning framework for DLMs. The core idea is to replace the standard random masking strategy with a trajectory-aware masking strategy that injects trajectory information into the training process. Instead of collecting trajectory signals from costly diffusion sampling, \method{} uses a much lighter alternative: an autoregressive teacher estimates token-level difficulty with a single teacher-forcing pass over the training data. We then discretize these difficulty scores into buckets and construct a \emph{hard-to-easy} masking strategy. This design encourages the model to learn a hard-to-easy decoding order that improves parallelism while keeping the training pipeline simple and unchanged from standard MDLM training.

We evaluate \method{} on two open-source DLMs, LLaDA-Instruct and Dream-Instruct, across four widely used math and coding benchmarks. The results show that \method{} achieves up to $3\times$ higher parallelism while maintaining similar accuracy compared to train-free acceleration baselines and standard MDLM training. Moreover, \method{} introduces trajectory supervision with almost no additional cost over standard MDLM training, requiring only 3 hours on $8\times$ A100 GPUs, while achieving performance comparable to distillation-based methods, which typically require costly diffusion sampling to generate separate distillation datasets for each target model. Ablation studies confirm the contribution of each component, and trajectory analysis shows that models trained with \method{} exhibit more efficient decoding trajectories, validating the effectiveness of trajectory-guided supervision.

Our main contributions are as follows:
\begin{itemize}
    \item We propose a trajectory-guided SFT framework that addresses the train--inference mismatch in DLMs by using token-level difficulty from an AR teacher to construct trajectory-aware supervision.
    \item We introduce a trajectory-aware masking strategy that simulates hard-to-easy decoding with minimal changes to standard MDLM training.
    \item We show that \method{} achieves competitive performance with much lower overhead than prior trajectory optimization methods, requiring only a single AR forward pass and no costly diffusion sampling.
\end{itemize}

\begin{figure*}[t]
  \centering
  \includegraphics[width=\linewidth]{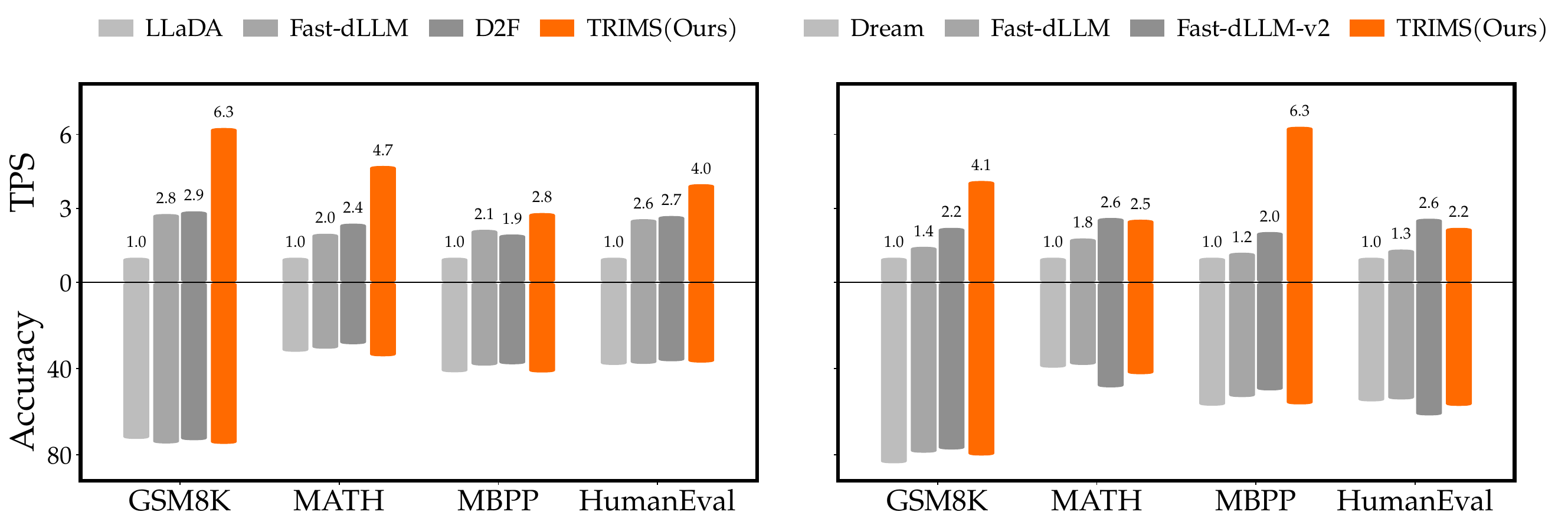}
  \caption{\method{} improves the accuracy-parallelism trade-off on both LLaDA-Instruct (left) and Dream-Instruct (right), achieving higher TPS (tokens predicted per step) while maintaining competitive accuracy, with much lower training cost than distillation-based methods.}  \label{fig:teaser}
\end{figure*}

\section{Preliminaries}

\paragraph{Masked Diffusion Language Models (MDLMs).}
Unlike autoregressive language models that generate tokens in a strict left-to-right order, masked diffusion language models (MDLMs) formulate generation as an iterative denoising process over partially masked sequences \citep{austin2021structured, shi2024simplified}. 
Given a clean sequence $x_0 = (x_0^1, \dots, x_0^L)$, the forward corruption process randomly replaces tokens with a special \texttt{[MASK]} symbol.
At corruption level $t \in [0,1]$, the corrupted sequence $x_t$ is sampled as:
\[
q(x_t \mid x_0) =
\prod_{i=1}^{L}
\left[(1-t)\,\mathbf{1}[x_t^i = x_0^i] + t\,\mathbf{1}[x_t^i = \texttt{[MASK]}] \right].
\]
Here, $\mathbf{1}[\cdot]$ denotes the indicator function.
The reverse model $p_\theta$ learns to recover tokens from the corrupted sequence.
At each step, the model predicts all masked tokens in parallel:
\[
p_\theta(x_0 \mid x_t) =
\prod_{i : x_t^i = \texttt{[MASK]}}
p_\theta(x_0^i \mid x_t).
\]
The training objective minimizes the negative log-likelihood over masked positions:
\[
\mathcal{L}(\theta) =
-\mathbb{E}_{t,x_0,x_t}
\left[
\sum_{i=1}^{L}
\mathbf{1}[x_t^i = \texttt{[MASK]}]
\log p_\theta(x_0^i \mid x_t)
\right].
\]

\paragraph{Iterative Decoding and Trajectory.}
During inference, MDLMs decode sequences through iterative refinement rather than a single one-shot prediction. At intermediate decoding steps, some tokens have already been resolved while others remain masked for later prediction. This induces an implicit token reveal order during generation, which we refer to as the decoding trajectory. Different trajectories can lead to different levels of token predictability in later steps, and therefore to different accuracy--parallelism trade-offs. Our method builds on this view by introducing training signals that encourage more effective reveal orders.
\section{Method}

\begin{figure}[t]
  \centering
  \includegraphics[width=\linewidth,trim={1.2cm 15.2cm 0.2cm 10.0cm},clip]{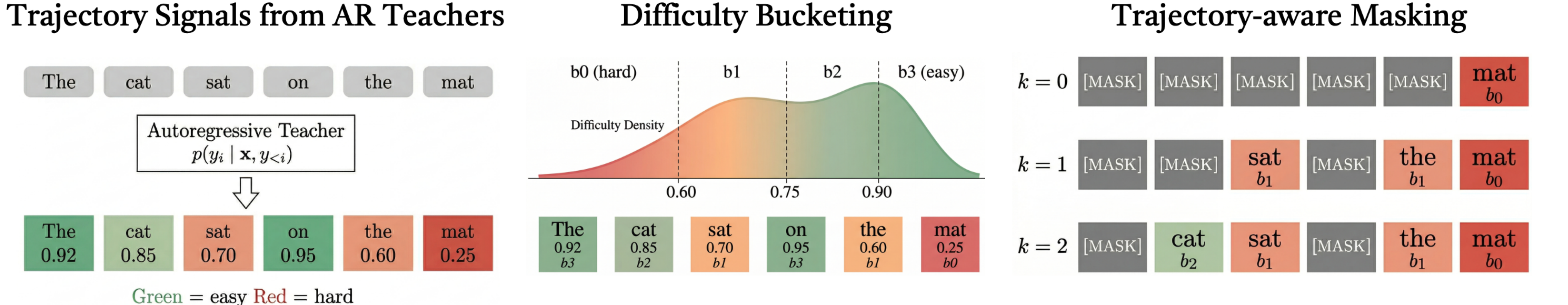}
  \caption{\method{} consists of three stages: (1) an AR teacher estimates token-level difficulty scores, (2) scores are discretized into ordered buckets, and (3) bucket assignments drive trajectory-aware masking that simulates a hard-to-easy decoding order during training.}
  \label{fig:method-overview}
\end{figure}

This section introduces the proposed \method{}. We first describe how we derive token-level trajectory signals from an autoregressive teacher, and then show how these signals are incorporated into training through a trajectory-aware masking strategy. This design injects trajectory supervision into standard MDLM training with minimal overhead and avoids costly diffusion sampling. Figure~\ref{fig:method-overview} illustrates the overall three-stage pipeline of \method{}.

\subsection{Trajectory Signals from AR Teachers}

To obtain trajectory supervision without expensive sampling, we construct token-level trajectory signals using an AR teacher model. 
Instead of explicitly observing decoding trajectories through model rollouts, we approximate token priority using prediction statistics from an AR model, which only requires a single forward pass over the training data.

Given a prompt--completion pair $(x, y)$, we run the AR teacher in teacher-forcing mode.
For each completion token $y_i$, the teacher provides the conditional probability $p_T(y_i \mid x, y_{<i})$, from which we compute the token difficulty score as:
\[
s_i = -\log p_T(y_i \mid x, y_{<i}),
\]
where larger values indicate harder tokens for the teacher to predict.

\begin{algorithm}[t]
    \caption{\textbf{\fullmethod{}}}
    \label{alg:tg-sft-training}
    \setlength{\baselineskip}{10pt}%
    \setlength{\parskip}{0.25ex}%
    \begin{algorithmic}[1]
    
    \Statex \texttt{// Differences from standard MDLM training are highlighted in \textcolor{brown}{brown}.}
    
    \State \textbf{Input:} Training data $\mathcal{D}$ with bucket ids $b_i$, model $x_\theta$ with parameters $\theta$, corruption process $q$, noise schedule $\alpha_t$, and masking probabilities $p_{\text{context}}$ and $p_{\text{future}}$
    
    \Repeat
        \State Sample $(\mathbf{x}^{1:L}, \mathbf{b}^{1:L})$ i.i.d. from $\mathcal{D}$
        \State Sample $t \sim \mathcal{U}[0,1]$
        \State Compute $\alpha_t, \dot{\alpha}_t$
        \State \textcolor{brown}{Sample bucket threshold $k \sim \mathcal{U}\{0, K-1\}$}
        \State Sample $\mathbf{z}_t^{1:L} \sim q(\mathbf{z}_t^{1:L} \mid \mathbf{x})$
        \For{$\ell = 1$ \textbf{to} $L$}
            \If{$b^\ell > k$}
                \State \textcolor{brown}{mask $z_t^\ell$ with probability $p_{\text{future}}$}
            \Else
                \State \textcolor{brown}{mask $z_t^\ell$ with probability $p_{\text{context}}$}
            \EndIf
        \EndFor
        \State Compute $x_\theta(\mathbf{z}_t^{1:L})$
    
        \State $\mathcal{L}^{\mathrm{MDM}}(\theta) \leftarrow
        \frac{\dot{\alpha}_t}{1-\alpha_t}
        \sum_{\ell=1}^{L}
        \delta_{z_t^\ell,m}
        \log \left\langle x_\theta^\ell(\mathbf{z}_t^{1:L}), x^\ell \right\rangle$
    
        \State Perform gradient descent step on $\mathcal{L}^{\mathrm{MDM}}(\theta)$
    
    \Until converged
    
    \State \textbf{Return} $\theta$
    
    \end{algorithmic}
\end{algorithm}

\subsection{Difficulty Bucketing}

Rather than using raw continuous difficulty scores, we discretize them into $K$ ordered buckets via quantile-based thresholds. This choice is motivated by two considerations. First, quantile-based thresholds are invariant to the absolute scale of NLL scores, making the method robust to score distribution shifts across datasets or AR teachers. Second, equal-mass quantiles ensure each bucket contains roughly the same number of tokens, keeping all trajectory stages balanced and preventing any single stage from dominating training.

Specifically, we collect difficulty scores of all completion tokens across the training data and compute quantile thresholds that divide the score distribution into $K$ equal-mass intervals. Each token is assigned a bucket id $b_i \in \{0, \dots, K-1\}$ by its quantile rank, where harder tokens (higher $s_i$) receive lower bucket ids, placing them into earlier decoding stages.

\subsection{Trajectory-aware Masking}

Using the bucket assignments, we construct a trajectory-aware masking strategy that simulates iterative decoding during training. Our key intuition is to \emph{resolve difficult tokens earlier in the decoding process}. This benefits parallel decoding in two ways: (1) once difficult tokens are determined, the remaining tokens often become more predictable, enabling more tokens to be generated in parallel; and (2) resolving difficult tokens early, while many positions remain masked, preserves sufficient flexibility for subsequent steps to refine the remaining tokens around these anchored positions.

Let $b_i$ denote the bucket id of token $y_i$, where smaller bucket ids correspond to harder tokens. During training, we first sample a bucket threshold $k$ uniformly from $\{0, \dots, K-1\}$. Tokens satisfying $b_i \le k$ are treated as already available context, while tokens with $b_i > k$ are treated as future tokens that should remain masked. To emphasize learning on these future tokens while maintaining robustness, we use different masking probabilities for the two groups.
Formally, the masking probability $m_i$ for token $y_i$ is defined as:
\[
m_i =
\begin{cases}
 p_{\text{context}} & \text{if } b_i \le k, \\
p_{\text{future}} & \text{if } b_i > k,
\end{cases},
\]
where $p_{\text{future}} = 0.95$ ensures future tokens are nearly always masked, and $p_{\text{context}} = 0.05$ introduces a small amount of masking on context tokens for robustness, preventing the model from overfitting to perfectly revealed context during training.

To reduce the mismatch with the pretraining stage and maintain training stability, we apply the trajectory-aware masking scheme to only $10\%$ of the training data. The remaining data follow the standard MDLM masking process, and we keep the original ELBO objective unchanged.
Algorithm~\ref{alg:tg-sft-training} summarizes the overall training procedure of \method{}.

\section{Experiments}

\subsection{Experimental Details}

In this section, we evaluate \method{} on two open-source DLMs and analyze its effectiveness, efficiency, and impact on decoding trajectories.

\begin{figure}[t]
  \setlength{\abovecaptionskip}{0.4em}%
  \centering
  \includegraphics[width=\linewidth]{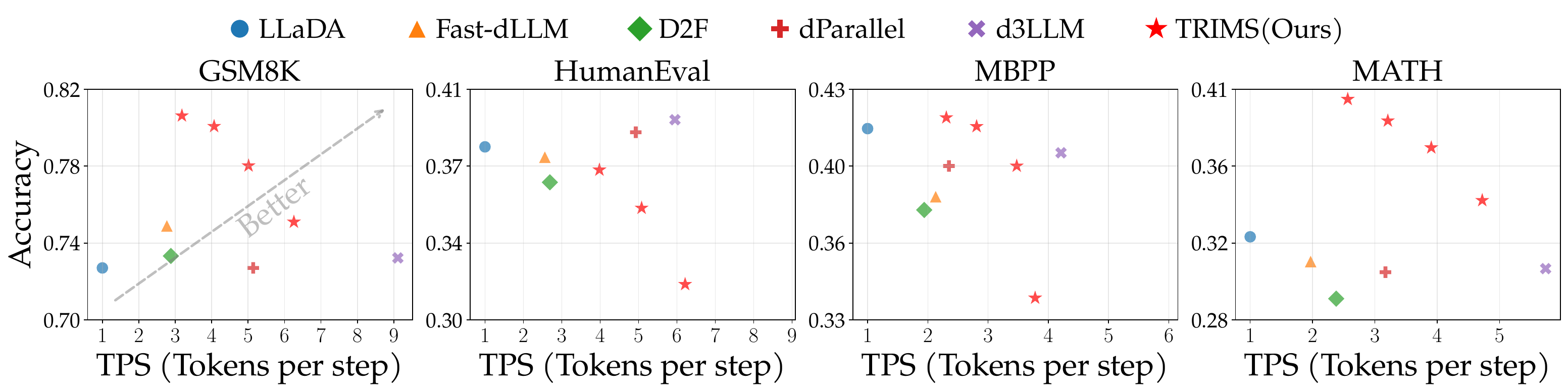}
  \caption{Accuracy-parallelism trade-off of \method{} and baselines on LLaDA-Instruct. The upper-right region indicates better performance with higher accuracy and parallelism.}
  \label{fig:llada-baseline}
\end{figure}
\vspace{-0.5em}
\begin{table}[t]
    \centering
    \small
    \setlength{\tabcolsep}{5pt}
    \begin{tabular}{lcccccccc}
    \toprule
    \multirow{2}{*}{Method} & \multicolumn{2}{c}{GSM8K} & \multicolumn{2}{c}{MATH} & \multicolumn{2}{c}{MBPP} & \multicolumn{2}{c}{HumanEval} \\
    \cmidrule(lr){2-3} \cmidrule(lr){4-5} \cmidrule(lr){6-7} \cmidrule(lr){8-9}
    & Acc$(\uparrow)$ & TPS$(\uparrow)$ & Acc$(\uparrow)$ & TPS$(\uparrow)$ & Acc$(\uparrow)$ & TPS$(\uparrow)$ & Acc$(\uparrow)$ & TPS$(\uparrow)$ \\
    \midrule
    LLaDA & $72.6$ & $1.00$ & $\underline{32.2}$ & $1.00$ & $\underline{41.7}$ & $1.00$ & $38.3$ & $1.00$ \\
    Fast-dLLM & $\underline{74.7}$ & $2.77$ & $30.8$ & $1.97$ & $38.6$ & $2.13$ & $37.8$ & $2.56$ \\
    D2F & $73.2$ & $2.88$ & $28.7$ & $2.38$ & $38.0$ & $1.94$ & $36.6$ & $2.69$ \\
    dParallel & $72.6$ & $\underline{5.14}$ & $30.2$ & $3.17$ & $40.0$ & $2.35$ & $\underline{39.0}$ & $\underline{4.93}$ \\
    d3LLM & $73.1$ & $\textbf{9.11}$ & $30.4$ & $\textbf{5.74}$ & $40.6$ & $\textbf{4.21}$ & $\textbf{39.6}$ & $\textbf{5.95}$ \\
    \rowcolor{gray!15}
    \method{} & $\textbf{74.9}$ & $\underline{6.26}$ & $\textbf{34.3}$ & $\underline{4.72}$ & $\textbf{41.8}$ & $\underline{2.81}$ & $37.2$ & $3.98$ \\
    \bottomrule
    \end{tabular}
    \vspace{-0.5em}
    \caption{Comparison of \method{} with baseline methods on LLaDA-Instruct. TPS denotes tokens per step and higher values indicate better parallelism.}
    \label{tab:main-results-llada}
    \end{table}

\paragraph{Dataset Preparation.}
We use the s1K dataset~\citep{muennighoff2025s1} for training. Reasoning data typically produces longer completions, making it a natural choice for evaluating the parallel decoding behavior of DLMs.
For the AR teacher, we use Qwen3-8B~\citep{yang2025qwen3} to compute the token-level difficulty scores.

\paragraph{Model and Training Details.}
In all experiments, we use LLaDA-Instruct~\citep{nielarge} and Dream-Instruct~\citep{ye2025dream} as the backbone models and implement \method{} in the dLLM codebase~\citep{zhou2026dllmsimplediffusionlanguage}. We use a sequence length of $1,024$, train for $30$ epochs with a batch size $32$ and the gradient accumulation factor $4$, and set the learning rate at $1\times10^{-4}$. The loss is computed only on response tokens. We apply LoRA~\citep{hu2022lora} with rank $r=32$, scaling factor $\alpha=32$, and the weight decay $0.1$. We use a cosine learning rate schedule with warmup ratio $0.03$. Training is conducted on $8\times$NVIDIA A100 80GB GPUs with DeepSpeed ZeRO-2~\citep{rajbhandari2020zero} enabled.

\paragraph{Baselines.}
To evaluate the effectiveness of \method{}, we compare it with four groups of baselines: (1) the original DLM backbones, LLaDA-Instruct~\citep{nielarge} and Dream-Instruct~\citep{ye2025dream}; (2) train-free acceleration methods, Fast-dLLM~\citep{wu2025fastdllm}; (3) training-based acceleration methods, Fast-dLLM-v2~\citep{wu2025fastdllmv2} and D2F~\citep{wang2025diffusionllmsfasterthanarinference}; and (4) distillation-based methods, dParallel~\citep{chen2025dparallel} and d3LLM~\citep{qian2026d3llm}. For evaluation, we report results on four widely used benchmarks: GSM8K~\citep{cobbe2021gsm8k}, MATH~\citep{hendrycks2021math}, HumanEval~\citep{chen2021humaneval}, and MBPP~\citep{austin2021mbpp}. These benchmarks assess core mathematical and coding abilities, and their relatively long solution sequences make them suitable for evaluating parallel decoding. We report both accuracy and parallelism, where parallelism is measured by tokens predicted per step (TPS). \method{} uses the confidence-based sampling strategy from Fast-dLLM~\citep{wu2025fastdllm}.

\subsection{Main Results}
\paragraph{LLaDA Results.}
As shown in Table~\ref{tab:main-results-llada}, \method{} achieves strong parallelism gains across benchmarks, reaching $6.26$ TPS on GSM8K and $4.72$ TPS on MATH while maintaining competitive accuracy. The speedup on coding benchmarks is more modest, since the rigid syntactic structure of code makes tokens more predictable, leaving less room for trajectory supervision to improve parallel decoding.
Compared with train-free baselines, \method{} consistently improves parallelism while preserving accuracy.
Compared with distillation-based methods, \method{} achieves competitive results on the accuracy-parallelism trade-off. We note that d3LLM achieves higher raw TPS on some benchmarks; however, as shown in Table~\ref{tab:training-overhead}, this comes at the cost of 287 GPU hours of curation compute per target model and 93K training examples, nearly two orders of magnitude more data than \method{}'s 1K samples. The TPS gap therefore reflects a deliberate resource-efficiency trade-off rather than a fundamental capability difference.
Figure~\ref{fig:llada-baseline} shows the full accuracy-parallelism trade-off curve. The frontier of \method{} lies mostly in the upper-right region, indicating a favorable balance between the accuracy and parallelism relative to all baselines.

\paragraph{Dream Results.} 
\begin{figure}[t]
  \setlength{\abovecaptionskip}{0.4em}
  \centering
  \includegraphics[width=\linewidth]{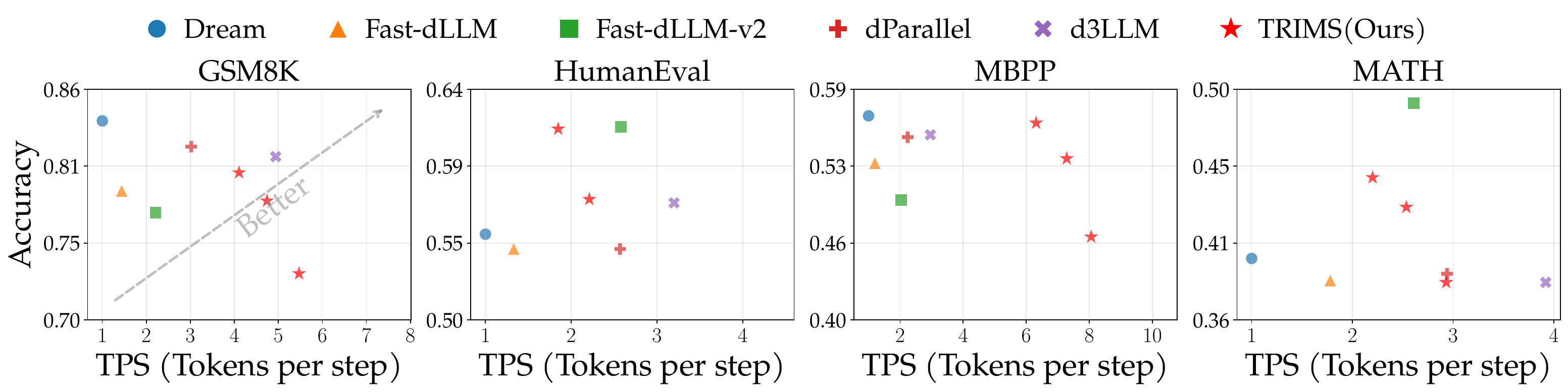}
  \caption{Accuracy-parallelism trade-off of \method{} and baselines on Dream-Instruct.}
  \label{fig:dream-baseline}
\end{figure}
\begin{table}[t]
    \centering
    \small
    \setlength{\tabcolsep}{5pt}
    \begin{tabular}{lcccccccc}
    \toprule
    \multirow{2}{*}{Method} & \multicolumn{2}{c}{GSM8K} & \multicolumn{2}{c}{MATH} & \multicolumn{2}{c}{MBPP} & \multicolumn{2}{c}{HumanEval} \\
    \cmidrule(lr){2-3} \cmidrule(lr){4-5} \cmidrule(lr){6-7} \cmidrule(lr){8-9}
    & Acc$(\uparrow)$ & TPS$(\uparrow)$ & Acc$(\uparrow)$ & TPS$(\uparrow)$ & Acc$(\uparrow)$ & TPS$(\uparrow)$ & Acc$(\uparrow)$ & TPS$(\uparrow)$ \\
    \midrule
    Dream & $\textbf{83.9}$ & $1.00$ & $39.6$ & $1.00$ & $\textbf{57.2}$ & $1.00$ & $55.2$ & $1.00$ \\
    Fast-dLLM & $79.0$ & $1.44$ & $38.3$ & $1.78$ & $53.2$ & $1.20$ & $54.3$ & $1.33$ \\
    Fast-dLLM-v2 & $77.5$ & $2.21$ & $\textbf{48.7}$ & $2.61$ & $50.1$ & $2.04$ & $\textbf{61.7}$ & $\underline{2.58}$ \\
    dParallel & $\underline{82.1}$ & $\underline{3.02}$ & $38.7$ & $\underline{2.94}$ & $55.4$ & $2.24$ & $54.3$ & $2.57$ \\
    d3LLM & $81.4$ & $\textbf{4.94}$ & $38.2$ & $\textbf{3.92}$ & $55.6$ & $\underline{2.96}$ & $57.1$ & $\textbf{3.20}$ \\
    \rowcolor{gray!15}
    \method{} & $80.3$ & $\underline{4.11}$ & $\underline{42.6}$ & $2.54$ & $\underline{56.6}$ & $\textbf{6.31}$ & $\underline{57.3}$ & $2.21$ \\
    \bottomrule
    \end{tabular}
    \vspace{-0.5em}
    \caption{Comparison of \method{} with baseline methods on Dream-Instruct. TPS denotes tokens per step and higher values indicate better parallelism.}
    \label{tab:main-results-dream}
    \end{table}

As shown in Table~\ref{tab:main-results-dream}, \method{} achieves notable parallelism gains across benchmarks, with a particularly strong result on MBPP at $6.31$ TPS, corresponding to up to $3\times$ higher parallelism than the train-free baselines. One possible reason is that MBPP often contains relatively long prompts and test cases, where the trajectory learned by \method{} can be more fully exploited.
Compared with train-free baselines, \method{} consistently improves parallelism while maintaining similar accuracy.
Compared with distillation-based methods, \method{} achieves competitive results on the accuracy-parallelism trade-off; any remaining TPS gap with d3LLM reflects the same resource-efficiency difference discussed above.
Overall, the trend on Dream-Instruct is slightly different from that on LLaDA-Instruct. We hypothesize that this difference comes from Dream-Instruct being adapted from an autoregressive model~\citep{gong2024scaling}, which may lead to different behavior after diffusion post-training.
Figure~\ref{fig:dream-baseline} shows the full accuracy-parallelism trade-off curve. The frontier of \method{} lies mostly in the upper-right region, indicating a favorable balance between the accuracy and parallelism relative to all baselines.

\begin{table}[th]
    \centering
    \small
    \begin{tabular}{lcccc}
    \toprule
    Method & Supervision & Curation Compute & Train Data & Epochs \\
    \midrule
    \method{} & AR teacher & 0.6 GPU Hours & 1K & 30 \\
    dParallel & DLM distillation & 287 GPU Hours * N(model) & 93K & 6 \\
    d3LLM & DLM distillation & 287 GPU Hours * N(model) & 93K & 6 \\
    \bottomrule
    \end{tabular}
    \caption{Training overhead comparison. All GPU hours are measured on NVIDIA A100-80GB. Distillation-based methods require expensive DLM sampling for dataset curation and are trained on substantially larger datasets. \method{} needs almost no data curation cost with far less training data requirements.}
    \label{tab:training-overhead}
    \end{table}
\vspace{-0.5em}

\paragraph{Training Overhead.}
Table~\ref{tab:training-overhead} compares the training cost of \method{} with distillation-based methods. The cost of \method{} is close to standard MDLM training, requiring only a single teacher-forcing pass from an AR model over the training data. By contrast, distillation-based methods such as d3LLM and dParallel require additional diffusion sampling to construct distillation datasets. Unlike autoregressive decoding, DLM sampling relies on bidirectional attention and lacks full KV cache support and related low-level optimizations, leading to substantially higher time complexity per sampling step. Moreover, these methods typically need to construct a separate distillation dataset for each target model. In our setting, \method{} uses only 1K training examples versus 93K for distillation-based methods, nearly two orders of magnitude less curated data, further highlighting the data efficiency of \method{}.
Details of the training cost calculation are provided in Appendix~\ref{appendix:training-cost}.

\begin{figure}[t]
  \centering
  \begin{subfigure}[t]{0.95\linewidth}
    \centering
    \includegraphics[width=\linewidth]{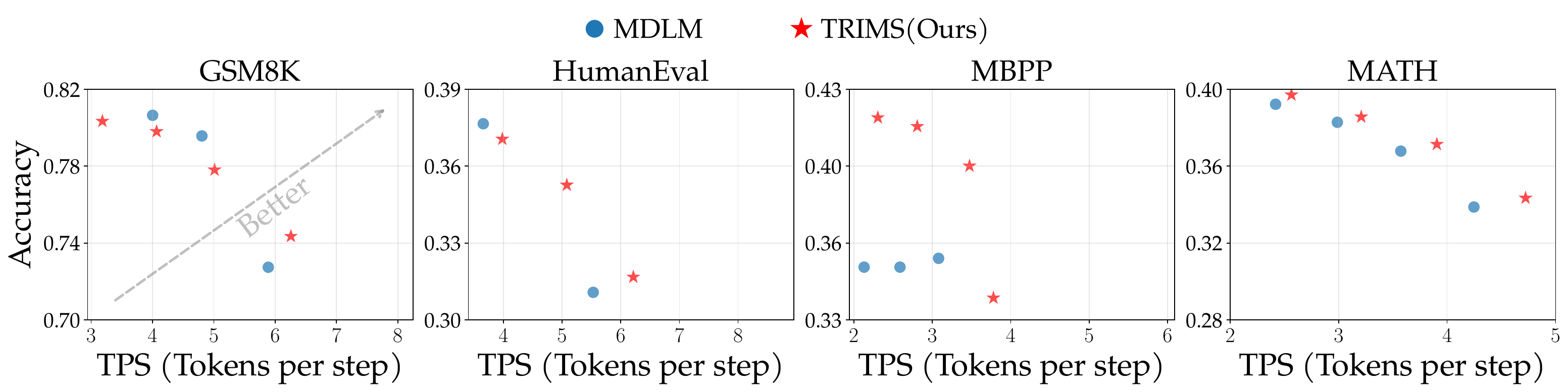}
    \caption{LLaDA.}
    \label{fig:ablation-trajectory-llada}
  \end{subfigure}

  \begin{subfigure}[t]{0.95\linewidth}
    \centering
    \includegraphics[width=\linewidth,trim={0 0 0 1.5cm},clip]{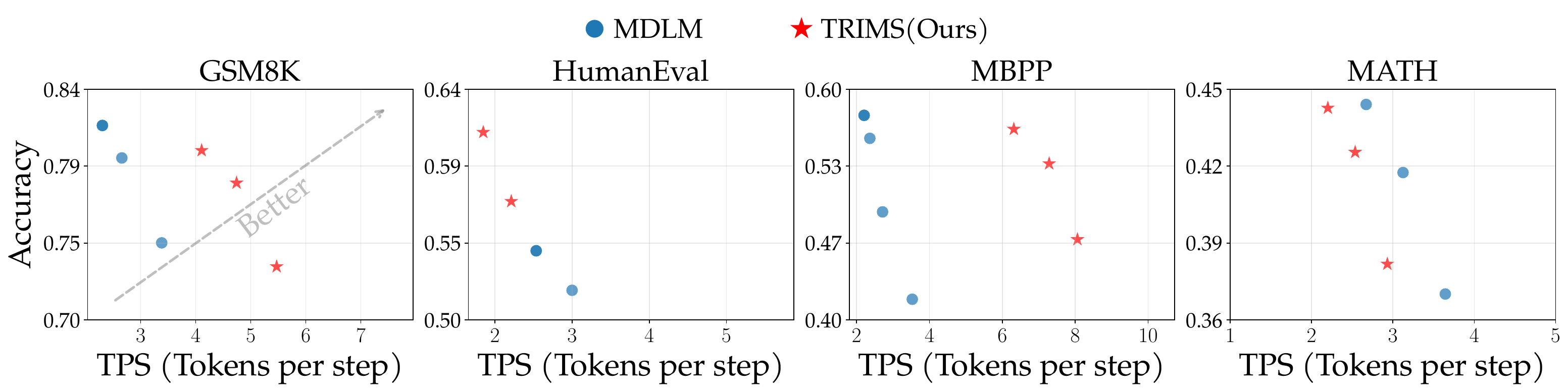}
    \caption{Dream.}
    \label{fig:ablation-trajectory-dream}
  \end{subfigure}
  \caption{Effect of trajectory supervision compared with standard MDLM training. \method{} consistently improves the accuracy-parallelism trade-off on most benchmarks, with especially clear gains on coding tasks.}
  \label{fig:ablation-trajectory}
\end{figure}
\vspace{-0.5em}

\subsection{Ablation Studies}
We conduct ablation studies to isolate the contribution of each component in \method{}, covering trajectory supervision, difficulty-based token ordering, bucket-based discretization, and difficulty metrics. Unless otherwise specified, all ablation experiments use the same training and evaluation settings as the main results.

\paragraph{Effect of Trajectory Supervision.}
We first evaluate whether introducing trajectory supervision improves performance over standard MDLM training. 
Figure~\ref{fig:ablation-trajectory} compares \method{} with standard MDLM training on both LLaDA and Dream. \method{} consistently outperforms the standard training setup on most benchmarks, showing that explicit trajectory supervision indeed improves the accuracy-parallelism trade-off. The gains are especially clear on coding benchmarks, including HumanEval and MBPP, where \method{} achieves both higher accuracy and stronger parallelism. On LLaDA, the improvement is consistent across all benchmarks. On Dream, MATH shows similar accuracy but slightly lower parallelism, possibly because Dream-Instruct is adapted from an AR model already well-trained on mathematical tasks, making trajectory post-training slightly out-of-distribution for this domain.

\paragraph{Effect of Trajectory Ordering.}
We compare three strategies for assigning tokens to buckets: difficulty-descending (harder tokens in earlier buckets), difficulty-ascending (easier tokens first), and random (uniform bucket assignment).
\begin{figure}[t]
  \centering
  \begin{subfigure}[t]{0.95\linewidth}
    \centering
    \includegraphics[width=\linewidth]{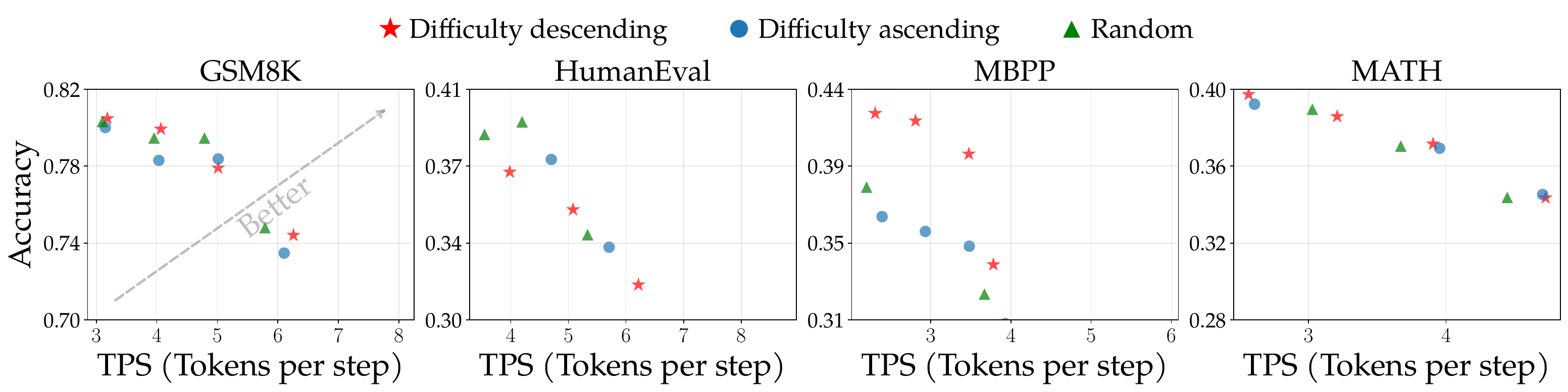}
    \caption{LLaDA.}
    \label{fig:ablation-ordering-llada}
  \end{subfigure}

  \begin{subfigure}[t]{0.95\linewidth}
    \centering
    \includegraphics[width=\linewidth,trim={0 0 0 1.6cm},clip]{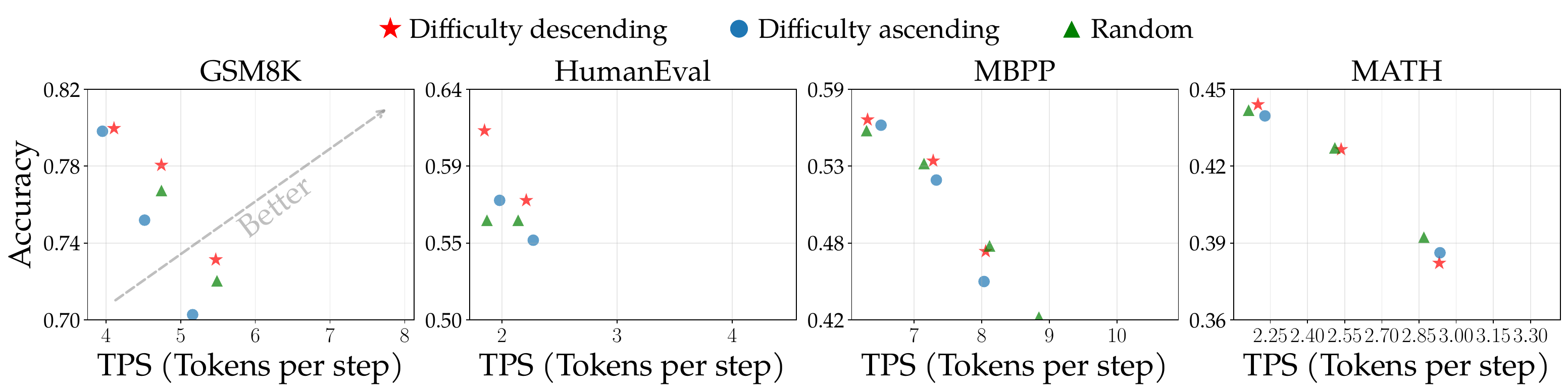}
    \caption{Dream.}
    \label{fig:ablation-ordering-dream}
  \end{subfigure}
  \caption{Effect of trajectory ordering under random, hard-to-easy, and easy-to-hard supervision. Hard-to-easy achieves the best overall performance, while all orderings improve over standard MDLM training.}
  \label{fig:ablation-ordering}
\end{figure}
Figure~\ref{fig:ablation-ordering} shows the results of the trajectory-ordering comparison. The difficulty-descending setting achieves the best overall performance on most benchmarks, although the margin over the other two settings is sometimes modest. Notably, comparing Figure~\ref{fig:ablation-trajectory} and Figure~\ref{fig:ablation-ordering}, even random ordering improves over standard MDLM training on most benchmarks, indicating that the bucket-based masking structure itself provides value through trajectory-aware supervision, regardless of ordering. The difficulty-descending ordering yields an additional boost by providing a more informative training signal that aligns with the natural hard-to-easy decoding pattern. This decomposition is also consistent with concurrent work~\citep{ma2026maskdllmneedsmasked}.

\paragraph{Effect of Bucket Granularity.}
We next study another core module in the method: the bucket-based masking mechanism. 
We vary the number of difficulty buckets $K \in \{4, 8, 16\}$. As shown in Figure~\ref{fig:ablation-bucket} (Appendix~\ref{appendix:bucket-granularity}), $K=8$ achieves the best performance, with $K=4$ and $K=16$ being slightly worse. This is possibly because too many buckets lead to overly fine-grained trajectory stages that are hard for the model to distinguish, while too few buckets merge tokens of meaningfully different difficulty into the same stage, losing discriminative signal. Overall, the performance difference across bucket numbers is modest, indicating robustness to this hyperparameter.

\paragraph{Effect of Difficulty Metrics.}
We compare negative log-likelihood (NLL) and entropy as difficulty metrics from the autoregressive teacher.
As shown in Figure~\ref{fig:ablation-metric} (Appendix~\ref{appendix:difficulty-metrics}), the two metrics produce comparable results on mathematical benchmarks, while NLL yields a noticeably better accuracy-parallelism trade-off on coding tasks (HumanEval and MBPP). A possible explanation is that coding tasks have more rigid syntactic structure, where NLL's token-specific confidence signal can more precisely distinguish structurally critical tokens from predictable ones, leading to a more informative bucket assignment. Overall, the difference remains modest, indicating that \method{} is robust to the choice of difficulty metric.

\begin{figure}[ht]
  \centering
  \includegraphics[width=0.8\linewidth]{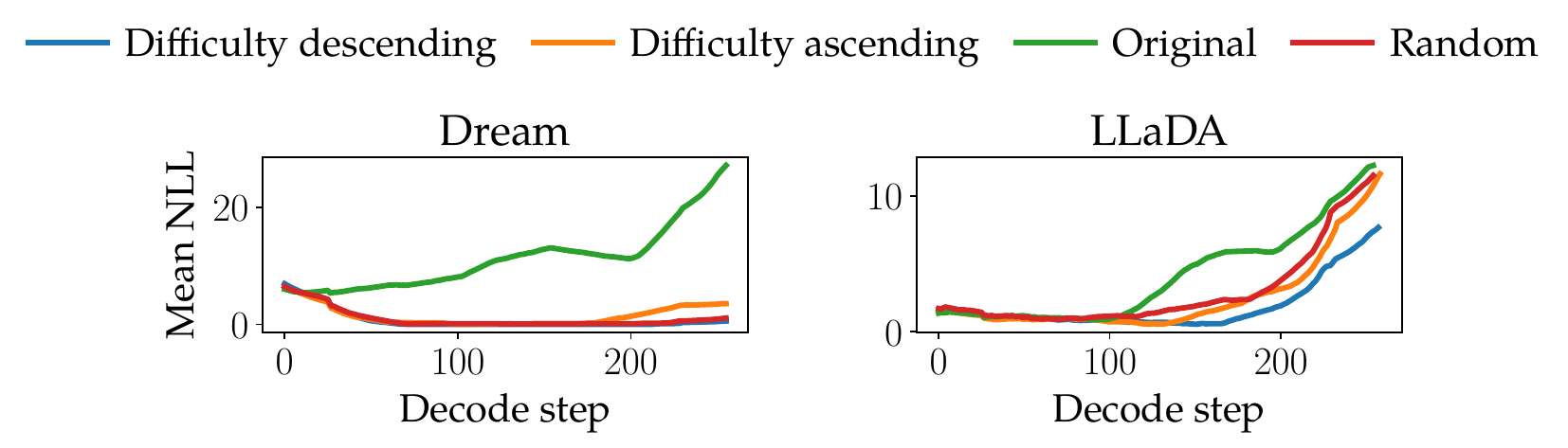}
  \caption{Per-step token NLL during decoding across \method{} variants and original DLM baselines, showing how uncertainty evolves over the generation process.}
  \label{fig:ablation-difficulty}
\end{figure}

\subsection{Trajectory Analysis}
To better understand how \method{} improves the accuracy-parallelism trade-off, we analyze the decoding trajectories of the trained models. Specifically, we study whether the learned trajectories reduce token uncertainty more efficiently during sampling.

\paragraph{Implementation Details.}
We use Qwen3-8B~\citep{yang2025qwen3} to compute the token-level NLL of generated sequences at each decoding step, and we evaluate on 40 randomly sampled examples from GSM8K~\citep{cobbe2021gsm8k}. We compare four settings: the original DLM models (LLaDA-Instruct~\citep{nielarge} and Dream-Instruct~\citep{ye2025dream}) and models trained with three \method{} masking strategies: difficulty-descending, difficulty-ascending, and random ordering. We set \texttt{max\_new\_tokens} and \texttt{steps} to $256$, and keep all other generation hyperparameters the same as in the main evaluation.

\paragraph{Results.}
As shown in Figure~\ref{fig:ablation-difficulty}, all three \method{} variants exhibit consistently lower NLL than the original models throughout decoding, except at the very early steps where all settings start from similar high-uncertainty states. This gap indicates that \method{} reduces token uncertainty faster during decoding, leading to more efficient trajectories. Among the variants, the difficulty-descending setting yields the lowest NLL across most steps, consistent with our design intuition: by resolving difficult tokens first, the model anchors high-uncertainty positions early, making the remaining tokens easier to predict in parallel at later steps.

\section{Conclusion}

In this paper, we present \fullmethod{} (\method{}), a simple yet effective trajectory guided supervised fine tuning framework for DLMs that mitigates the train inference mismatch by incorporating token level difficulty signals and a trajectory aware masking strategy. Experiments demonstrate that \method{} consistently improves the accuracy and parallelism trade off over standard MDLM training and acceleration baselines, achieving performance competitive with distillation based methods while requiring substantially lower training cost and data. Our results further suggest that lightweight trajectory supervision leads to more effective decoding behaviors, highlighting its potential as a practical and scalable approach for enhancing parallel decoding in DLMs. Future work includes extending this framework to larger scale models and exploring richer forms of trajectory signals.

\clearpage
\bibliography{colm2026_conference}

@misc{zhou2026dllmsimplediffusionlanguage,
      title={dLLM: Simple Diffusion Language Modeling}, 
      author={Zhanhui Zhou and Lingjie Chen and Hanghang Tong and Dawn Song},
      year={2026},
      eprint={2602.22661},
      archivePrefix={arXiv},
      primaryClass={cs.CL},
      url={https://arxiv.org/abs/2602.22661}, 
}

@misc{labs2025mercuryultrafastlanguagemodels,
      title={Mercury: Ultra-Fast Language Models Based on Diffusion}, 
      author={Inception Labs and Samar Khanna and Siddhant Kharbanda and Shufan Li and Harshit Varma and Eric Wang and Sawyer Birnbaum and Ziyang Luo and Yanis Miraoui and Akash Palrecha and Stefano Ermon and Aditya Grover and Volodymyr Kuleshov},
      year={2025},
      eprint={2506.17298},
      archivePrefix={arXiv},
      primaryClass={cs.CL},
      url={https://arxiv.org/abs/2506.17298}, 
}

@article{fu2025bits,
  title={From bits to rounds: Parallel decoding with exploration for diffusion language models},
  author={Fu, Hengyu and Huang, Baihe and Adams, Virginia and Wang, Charles and Srinivasan, Venkat and Jiao, Jiantao},
  journal={arXiv preprint arXiv:2511.21103},
  year={2025}
}

@article{ni2026flexibility,
  title={The Flexibility Trap: Why Arbitrary Order Limits Reasoning Potential in Diffusion Language Models},
  author={Ni, Zanlin and Wang, Shenzhi and Yue, Yang and Yu, Tianyu and Zhao, Weilin and Hua, Yeguo and Chen, Tianyi and Song, Jun and Yu, Cheng and Zheng, Bo and others},
  journal={arXiv preprint arXiv:2601.15165},
  year={2026}
}

@misc{ma2026maskdllmneedsmasked,
      title={Mask Is What DLLM Needs: A Masked Data Training Paradigm for Diffusion LLMs}, 
      author={Linrui Ma and Yufei Cui and Kai Han and Yunhe Wang},
      year={2026},
      eprint={2603.15803},
      archivePrefix={arXiv},
      primaryClass={cs.LG},
      url={https://arxiv.org/abs/2603.15803}, 
}

@article{chen2025dparallel,
  title={dparallel: Learnable parallel decoding for dllms},
  author={Chen, Zigeng and Fang, Gongfan and Ma, Xinyin and Yu, Ruonan and Wang, Xinchao},
  journal={arXiv preprint arXiv:2509.26488},
  year={2025}
}

@article{qian2026d3llm,
  title={d3LLM: Ultra-Fast Diffusion LLM using Pseudo-Trajectory Distillation},
  author={Qian, Yu-Yang and Su, Junda and Hu, Lanxiang and Zhang, Peiyuan and Deng, Zhijie and Zhao, Peng and Zhang, Hao},
  journal={arXiv preprint arXiv:2601.07568},
  year={2026}
}

@misc{wang2025diffusionllmsfasterthanarinference,
      title={Diffusion LLMs Can Do Faster-Than-AR Inference via Discrete Diffusion Forcing}, 
      author={Xu Wang and Chenkai Xu and Yijie Jin and Jiachun Jin and Hao Zhang and Zhijie Deng},
      year={2025},
      eprint={2508.09192},
      archivePrefix={arXiv},
      primaryClass={cs.LG},
      url={https://arxiv.org/abs/2508.09192}, 
}

@inproceedings{sohl2015deep,
  author    = {Sohl{-}Dickstein, Jascha and Weiss, Eric A. and Maheswaranathan, Niru and Ganguli, Surya},
  title     = {Deep Unsupervised Learning using Nonequilibrium Thermodynamics},
  booktitle = {Proceedings of the 32nd International Conference on Machine Learning},
  year      = {2015},
}

@inproceedings{ho2020denoising,
  author    = {Ho, Jonathan and Jain, Ajay and Abbeel, Pieter},
  title     = {Denoising Diffusion Probabilistic Models},
  booktitle = {Advances in Neural Information Processing Systems},
  year      = {2020}
}

@inproceedings{song2020score,
  author    = {Song, Yang and Sohl{-}Dickstein, Jascha and Kingma, Diederik P. and Kumar, Abhishek and Ermon, Stefano and Poole, Ben},
  title     = {Score-Based Generative Modeling through Stochastic Differential Equations},
  booktitle = {International Conference on Learning Representations},
  year      = {2021}
}

@inproceedings{austin2021structured,
  author    = {Austin, Jacob and Johnson, Daniel D. and Ho, Jonathan and Tarlow, Daniel and van den Berg, Rianne},
  title     = {Structured Denoising Diffusion Models in Discrete State-Spaces},
  booktitle = {Advances in Neural Information Processing Systems},
  year      = {2021}
}

@inproceedings{hoogeboom2021argmax,
  author    = {Hoogeboom, Emiel and Nielsen, Didrik and Jaini, Priyank and Forr{\'e}, Patrick and Welling, Max},
  title     = {Argmax Flows and Multinomial Diffusion: Learning Categorical Distributions},
  booktitle = {Advances in Neural Information Processing Systems},
  year      = {2021}
}

@inproceedings{campbell2022continuous,
  author    = {Campbell, Andrew and Benton, Joe and De Bortoli, Valentin and Rainforth, Thomas and Deligiannidis, George and Doucet, Arnaud},
  title     = {A Continuous Time Framework for Discrete Denoising Models},
  booktitle = {Advances in Neural Information Processing Systems},
  year      = {2022}
}

@inproceedings{sun2022score,
  author    = {Sun, Haoran and Yu, Lijun and Dai, Bo and Schuurmans, Dale and Dai, Hanjun},
  title     = {Score-based Continuous-time Discrete Diffusion Models},
  booktitle = {International Conference on Learning Representations},
  year      = {2023},
}

@inproceedings{meng2022concrete,
  author    = {Meng, Chenlin and Choi, Kristy and Song, Jiaming and Ermon, Stefano},
  title     = {Concrete Score Matching: Generalized Score Matching for Discrete Data},
  booktitle = {Advances in Neural Information Processing Systems},
  year      = {2022}
}

@inproceedings{lou2023discrete,
  author    = {Lou, Aaron and Meng, Chenlin and Ermon, Stefano},
  title     = {Discrete Diffusion Modeling by Estimating the Ratios of the Data Distribution},
  booktitle = {Proceedings of the 41st International Conference on Machine Learning},
  year      = {2024},
}

@inproceedings{sahoo2024simple,
  author    = {Sahoo, Subham Sekhar and Arriola, Marianne and Schiff, Yair and Gokaslan, Aaron and Marroquin, Edgar and Chiu, Justin T. and Rush, Alexander and Kuleshov, Volodymyr},
  title     = {Simple and Effective Masked Diffusion Language Models},
  booktitle = {Advances in Neural Information Processing Systems},
  year      = {2024}
}

@inproceedings{shi2024simplified,
  author    = {Shi, Jiaxin and Han, Kehang and Wang, Zhe and Doucet, Arnaud and Titsias, Michalis K.},
  title     = {Simplified and Generalized Masked Diffusion for Discrete Data},
  booktitle = {Advances in Neural Information Processing Systems},
  year      = {2024}
}

@inproceedings{ou2024your,
  author    = {Ou, Jingyang and Nie, Shen and Xue, Kaiwen and Zhu, Fengqi and Sun, Jiacheng and Li, Zhenguo and Li, Chongxuan},
  title     = {Your Absorbing Discrete Diffusion Secretly Models the Conditional Distributions of Clean Data},
  booktitle = {International Conference on Learning Representations},
  year      = {2025},
}

@inproceedings{zheng2024masked,
  author    = {Zheng, Kaiwen and Chen, Yongxin and Mao, Hanzi and Liu, Ming{-}Yu and Zhu, Jun and Zhang, Qinsheng},
  title     = {Masked Diffusion Models are Secretly Time-Agnostic Masked Models and Exploit Inaccurate Categorical Sampling},
  booktitle = {International Conference on Learning Representations},
  year      = {2025},
}

@inproceedings{li2022diffusion,
  author    = {Li, Xiang Lisa and Thickstun, John and Gulrajani, Ishaan and Liang, Percy and Hashimoto, Tatsunori B.},
  title     = {{Diffusion-LM} Improves Controllable Text Generation},
  booktitle = {Advances in Neural Information Processing Systems},
  year      = {2022}
}

@inproceedings{gong2022diffuseq,
  author    = {Gong, Shansan and Li, Mukai and Feng, Jiangtao and Wu, Zhiyong and Kong, Lingpeng},
  title     = {{DiffuSeq}: Sequence to Sequence Text Generation with Diffusion Models},
  booktitle = {International Conference on Learning Representations},
  year      = {2023}
}

@article{dieleman2022continuous,
  author    = {Dieleman, Sander and Sartran, Laurent and Roshannai, Arman and Savinov, Nikolay and Ganin, Yaroslav and Richemond, Pierre H. and Doucet, Arnaud and Strudel, Robin and Dyer, Chris and Durkan, Conor and others},
  title     = {Continuous Diffusion for Categorical Data},
  journal   = {arXiv preprint arXiv:2211.15089},
  year      = {2022}
}

@inproceedings{lin2023text,
  author    = {Lin, Zhenghao and Gong, Yeyun and Shen, Yelong and Wu, Tong and Fan, Zhihao and Lin, Chen and Duan, Nan and Chen, Weizhu},
  title     = {Text Generation with Diffusion Language Models: A Pre-Training Approach with Continuous Paragraph Denoise},
  booktitle = {Proceedings of the 40th International Conference on Machine Learning},
  year      = {2023},
}

@inproceedings{gat2024discrete,
  author    = {Gat, Itai and Remez, Tal and Shaul, Neta and Kreuk, Felix and Chen, Ricky T. Q. and Synnaeve, Gabriel and Adi, Yossi and Lipman, Yaron},
  title     = {Discrete Flow Matching},
  booktitle = {Advances in Neural Information Processing Systems},
  year      = {2024}
}

@inproceedings{havasi2025edit,
  author    = {Havasi, Marton and Karrer, Brian and Gat, Itai and Chen, Ricky T. Q.},
  title     = {Edit Flows: Variable Length Discrete Flow Matching with Sequence-Level Edit Operations},
  booktitle = {Advances in Neural Information Processing Systems},
  year      = {2025}
}

@article{nguyen2025oneflow,
  author    = {Nguyen, John and Havasi, Marton and Berrada, Tariq and Zettlemoyer, Luke and Chen, Ricky T. Q.},
  title     = {{OneFlow}: Concurrent Mixed-Modal and Interleaved Generation with Edit Flows},
  journal   = {arXiv preprint arXiv:2510.03506},
  year      = {2025}
}

@inproceedings{arriolablock,
  author    = {Arriola, Marianne and Gokaslan, Aaron and Chiu, Justin T. and Yang, Zhihan and Qi, Zhixuan and Han, Jiaqi and Sahoo, Subham Sekhar and Kuleshov, Volodymyr},
  title     = {Block Diffusion: Interpolating Between Autoregressive and Diffusion Language Models},
  booktitle = {International Conference on Learning Representations},
  year      = {2025},
}

@inproceedings{nielarge,
  author    = {Nie, Shen and Zhu, Fengqi and You, Zebin and Zhang, Xiaolu and Ou, Jingyang and Hu, Jun and Zhou, Jun and Lin, Yankai and Wen, Ji{-}Rong and Li, Chongxuan},
  title     = {Large Language Diffusion Models},
  booktitle = {Advances in Neural Information Processing Systems},
  year      = {2025},
}

@article{ye2025dream,
  author    = {Ye, Jiacheng and Xie, Zhihui and Zheng, Lin and Gao, Jiahui and Wu, Zirui and Jiang, Xin and Li, Zhenguo and Kong, Lingpeng},
  title     = {{Dream} {7B}: Diffusion Large Language Models},
  journal   = {arXiv preprint arXiv:2508.15487},
  year      = {2025}
}

@inproceedings{gong2024scaling,
  author    = {Gong, Shansan and Agarwal, Shivam and Zhang, Yizhe and Ye, Jiacheng and Zheng, Lin and Li, Mukai and An, Chenxin and Zhao, Peilin and Bi, Wei and Han, Jiawei and Peng, Hao and Kong, Lingpeng},
  title     = {Scaling Diffusion Language Models via Adaptation from Autoregressive Models},
  booktitle = {International Conference on Learning Representations},
  year      = {2025},
}

@inproceedings{horvitz2026rethinking,
  title     = {Rethinking Reasoning with Masked Diffusion Models},
  author    = {Horvitz, Zachary and Singhal, Raghav and Zou, Hao and Domingo-Enrich, Carles and Yu, Zhou and Ranganath, Rajesh and McKeown, Kathleen},
  booktitle = {International Conference on Learning Representations},
  year      = {2026},
  note      = {OpenReview: 2mkYB5KUwa}
}

@inproceedings{piskorz2026masks,
  title     = {Masks Can Be Distracting: On Context Comprehension in Diffusion Language Models},
  author    = {Piskorz, Julianna and Correia, Alvaro and Pinneri, Cristina and Alfarra, Motasem and Garrepalli, Risheek and Louizos, Christos},
  booktitle = {International Conference on Learning Representations},
  year      = {2026},
  note      = {OpenReview: CdJwNTisx1}
}

@article{li2026why,
  title   = {Why Diffusion Language Models Struggle with Truly Parallel (Non-Autoregressive) Decoding?},
  author  = {Li, Pengxiang and Muhtar, Dilxat and Yin, Lu and Chen, Tianlong and Liu, Shiwei},
  journal = {arXiv preprint arXiv:2602.23225},
  year    = {2026}
}

@article{yang2025taming,
  title   = {Taming Masked Diffusion Language Models via Consistency Trajectory Reinforcement Learning with Fewer Decoding Step},
  author  = {Yang, Jingyi and Chen, Guanxu and Hu, Xuhao and Shao, Jing},
  journal = {arXiv preprint arXiv:2509.23924},
  year    = {2025}
}

@inproceedings{wang2026spg,
  title     = {SPG: Sandwiched Policy Gradient for Masked Diffusion Language Models},
  author    = {Wang, Chenyu and Rashidinejad, Paria and Su, DiJia and Jiang, Song and Wang, Sid and Zhao, Siyan and Zhou, Cai and Shen, Shannon Zejiang and Chen, Feiyu and Jaakkola, Tommi and Tian, Yuandong and Liu, Bo},
  booktitle = {International Conference on Learning Representations},
  year      = {2026},
  note      = {OpenReview: 18j5Q49GwN}
}

@inproceedings{wu2025fastdllm,
  author    = {Wu, Chengyue and Zhang, Hao and Xue, Shuchen and Liu, Zhijian and Diao, Shizhe and Zhu, Ligeng and Luo, Ping and Han, Song and Xie, Enze},
  title     = {Fast-{dLLM}: Training-free Acceleration of Diffusion {LLM} by Enabling {KV} Cache and Parallel Decoding},
  booktitle = {International Conference on Learning Representations},
  year      = {2026}
}

@inproceedings{wu2025fastdllmv2,
  author    = {Wu, Chengyue and Zhang, Hao and Xue, Shuchen and Diao, Shizhe and Fu, Yonggan and Liu, Zhijian and Molchanov, Pavlo and Luo, Ping and Han, Song and Xie, Enze},
  title     = {Fast-{dLLM} v2: Efficient Block-Diffusion {LLM}},
  booktitle = {International Conference on Learning Representations},
  year      = {2026}
}

@inproceedings{ma2025dkv,
  author    = {Ma, Xinyin and Yu, Runpeng and Fang, Gongfan and Wang, Xinchao},
  title     = {{dKV-Cache}: The Cache for Diffusion Language Models},
  booktitle = {Advances in Neural Information Processing Systems},
  year      = {2025}
}

@inproceedings{ben2025accelerated,
  author    = {Ben{-}Hamu, Heli and Gat, Itai and Severo, Daniel and Nolte, Niklas and Karrer, Brian},
  title     = {Accelerated Sampling from Masked Diffusion Models via Entropy Bounded Unmasking},
  booktitle = {Advances in Neural Information Processing Systems},
  year      = {2025}
}

@inproceedings{christopher2025specdiff,
  author    = {Christopher, Jacob K. and Bartoldson, Brian R. and Ben{-}Nun, Tal and Cardei, Michael and Kailkhura, Bhavya and Fioretto, Ferdinando},
  title     = {Speculative Diffusion Decoding: Accelerating Language Generation through Diffusion},
  booktitle = {Proceedings of the 2025 Conference of the Nations of the Americas Chapter of the Association for Computational Linguistics: Human Language Technologies (Volume 1: Long Papers)},
  year      = {2025},
}

@inproceedings{israel2025planned,
  author    = {Israel, Daniel and Jin, Tian and Cheng, Ellie and Van den Broeck, Guy and Grover, Aditya and Subramanian, Suvinay and Carbin, Michael},
  title     = {Planned Diffusion},
  booktitle = {International Conference on Learning Representations},
  year      = {2026}
}

@article{liu2025tidar,
  author    = {Liu, Jingyu and Dong, Xin and Ye, Zhifan and Mehta, Rishabh and Fu, Yonggan and Singh, Vartika and Kautz, Jan and Zhang, Ce and Molchanov, Pavlo},
  title     = {{TiDAR}: Think in Diffusion, Talk in Autoregression},
  journal   = {arXiv preprint arXiv:2511.08923},
  year      = {2025}
}

@inproceedings{wang2025remasking,
  author    = {Wang, Guanghan and Schiff, Yair and Sahoo, Subham Sekhar and Kuleshov, Volodymyr},
  title     = {Remasking Discrete Diffusion Models with Inference-Time Scaling},
  booktitle = {Advances in Neural Information Processing Systems},
  year      = {2025}
}

@article{yang2025qwen3,
  title={Qwen3 technical report},
  author={Yang, An and Li, Anfeng and Yang, Baosong and Zhang, Beichen and Hui, Binyuan and Zheng, Bo and Yu, Bowen and Gao, Chang and Huang, Chengen and Lv, Chenxu and others},
  journal={arXiv preprint arXiv:2505.09388},
  year={2025}
}

@article{cobbe2021gsm8k,
  author    = {Cobbe, Karl and Kosaraju, Vineet and Bavarian, Mohammad and Chen, Mark and Jun, Heewoo and Kaiser, Lukasz and Plappert, Matthias and Tworek, Jerry and Hilton, Jacob and Nakano, Reiichiro and Hesse, Christopher and Schulman, John},
  title     = {Training Verifiers to Solve Math Word Problems},
  journal   = {arXiv preprint arXiv:2110.14168},
  year      = {2021}
}

@inproceedings{hendrycks2021math,
  author    = {Hendrycks, Dan and Burns, Collin and Kadavath, Saurav and Arora, Akul and Basart, Steven and Tang, Eric and Song, Dawn and Steinhardt, Jacob},
  title     = {Measuring Mathematical Problem Solving With the {MATH} Dataset},
  booktitle = {NeurIPS Datasets and Benchmarks},
  year      = {2021}
}

@article{chen2021humaneval,
  author    = {Chen, Mark and Tworek, Jerry and Jun, Heewoo and Yuan, Qiming and Pinto, Henrique Pond{\'{e}} de Oliveira and Kaplan, Jared and Edwards, Harri and Burda, Yuri and Joseph, Nicholas and Brockman, Greg and Ray, Alex and Puri, Raul and Krueger, Gretchen and Petrov, Michael and Khlaaf, Heidy and Sastry, Girish and Mishkin, Pamela and Chan, Brooke and Gray, Scott and Ryder, Nick and Pavlov, Mikhail and Power, Alethea and Kaiser, Lukasz and Bavarian, Mohammad and Winter, Clemens and Tillet, Philippe and Such, Felipe Petroski and Cummings, Dave and Plappert, Matthias and Chantzis, Fotios and Barnes, Elizabeth and Herbert{-}Voss, Ariel and Guss, William Hebgen and Nichol, Alex and Paino, Alex and Tezak, Nikolas and Tang, Jie and Babuschkin, Igor and Balaji, Suchir and Jain, Shantanu and Saunders, William and Hesse, Christopher and Carr, Andrew N. and Leike, Jan and Achiam, Josh and Misra, Vedant and Morikawa, Evan and Radford, Alec and Knight, Matthew and Brundage, Miles and Murati, Mira and Mayer, Katie and Welinder, Peter and McGrew, Bob and Amodei, Dario and McCandlish, Sam and Sutskever, Ilya and Zaremba, Wojciech},
  title     = {Evaluating Large Language Models Trained on Code},
  journal   = {arXiv preprint arXiv:2107.03374},
  year      = {2021}
}

@article{austin2021mbpp,
  author    = {Austin, Jacob and Odena, Augustus and Nye, Maxwell and Bosma, Maarten and Michalewski, Henryk and Dohan, David and Jiang, Ellen and Cai, Carrie and Terry, Michael and Le, Quoc and Sutton, Charles},
  title     = {Program Synthesis with Large Language Models},
  journal   = {arXiv preprint arXiv:2108.07732},
  year      = {2021}
}

@inproceedings{lightman2023let,
  title={Let's verify step by step},
  author={Lightman, Hunter and Kosaraju, Vineet and Burda, Yuri and Edwards, Harrison and Baker, Bowen and Lee, Teddy and Leike, Jan and Schulman, John and Sutskever, Ilya and Cobbe, Karl},
  booktitle={The twelfth international conference on learning representations},
  year={2024}
}

@article{li2024numinamath,
  title={Numinamath: The largest public dataset in ai4maths with 860k pairs of competition math problems and solutions},
  author={Li, Jia and Beeching, Edward and Tunstall, Lewis and Lipkin, Ben and Soletskyi, Roman and Huang, Shengyi and Rasul, Kashif and Yu, Longhui and Jiang, Albert Q and Shen, Ziju and others},
  journal={Hugging Face repository},
  volume={13},
  number={9},
  pages={9},
  year={2024}
}

@inproceedings{hu2022lora,
  author    = {Hu, Edward J. and Shen, Yelong and Wallis, Phillip and Allen-Zhu, Zeyuan and Li, Yuanzhi and Wang, Shean and Wang, Lu and Chen, Weizhu},
  title     = {{LoRA}: Low-Rank Adaptation of Large Language Models},
  booktitle = {International Conference on Learning Representations},
  year      = {2022}
}

@misc{zeng2025acecoderacingcoderrl,
      title={ACECODER: Acing Coder RL via Automated Test-Case Synthesis}, 
      author={Huaye Zeng and Dongfu Jiang and Haozhe Wang and Ping Nie and Xiaotong Chen and Wenhu Chen},
      year={2025},
      eprint={2502.01718},
      archivePrefix={arXiv},
      primaryClass={cs.SE},
      url={https://arxiv.org/abs/2502.01718}, 
}

@inproceedings{rajbhandari2020zero,
  author    = {Rajbhandari, Samyam and Rasley, Jeff and Ruwase, Olatunji and He, Yuxiong},
  title     = {{ZeRO}: memory optimizations toward training trillion parameter models},
  booktitle = {Proceedings of the International Conference for High Performance Computing, Networking, Storage and Analysis, SC 2020, Virtual Event / Atlanta, Georgia, USA, November 9-19, 2020},
  year      = {2020},
}

@inproceedings{muennighoff2025s1,
  author    = {Muennighoff, Niklas and Yang, Zitong and Shi, Weijia and Li, Xiang Lisa and Fei-Fei, Li and Hajishirzi, Hannaneh and Zettlemoyer, Luke and Liang, Percy and Cand{\`e}s, Emmanuel and Hashimoto, Tatsunori B},
  title     = {{s1}: Simple test-time scaling},
  booktitle = {Proceedings of the 2025 Conference on Empirical Methods in Natural Language Processing},
  year      = {2025},
}
\bibliographystyle{colm2026_conference}

\appendix
\clearpage
\section{Effect of Bucket Granularity}
\label{appendix:bucket-granularity}

\begin{figure}[ht]
  \centering
  \begin{subfigure}[t]{0.95\linewidth}
    \centering
    \includegraphics[width=\linewidth]{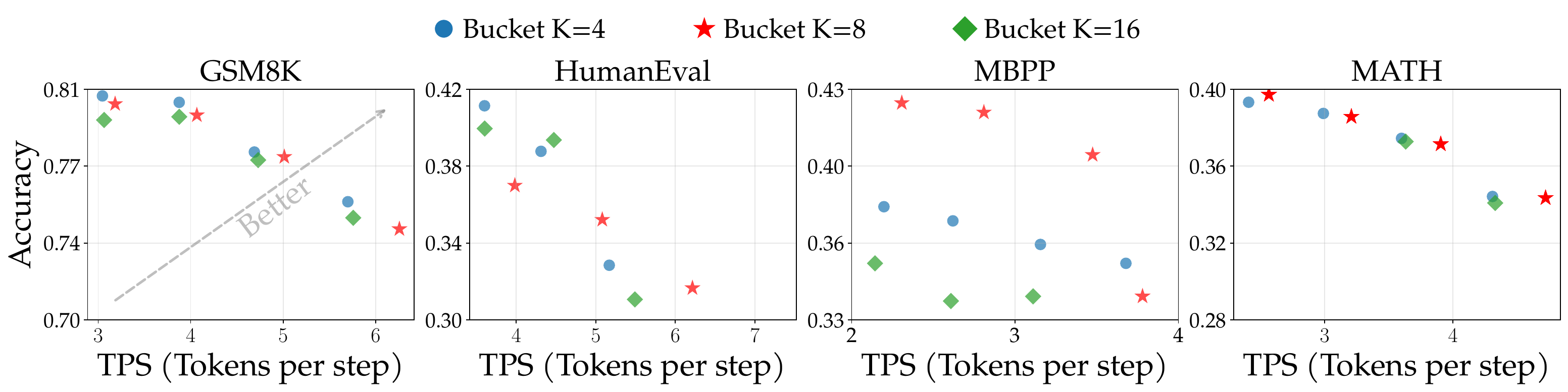}
    \caption{LLaDA.}
    \label{fig:ablation-bucket-llada}
  \end{subfigure}

  \begin{subfigure}[t]{0.95\linewidth}
    \centering
    \includegraphics[width=\linewidth,trim={0 0 0 1.5cm},clip]{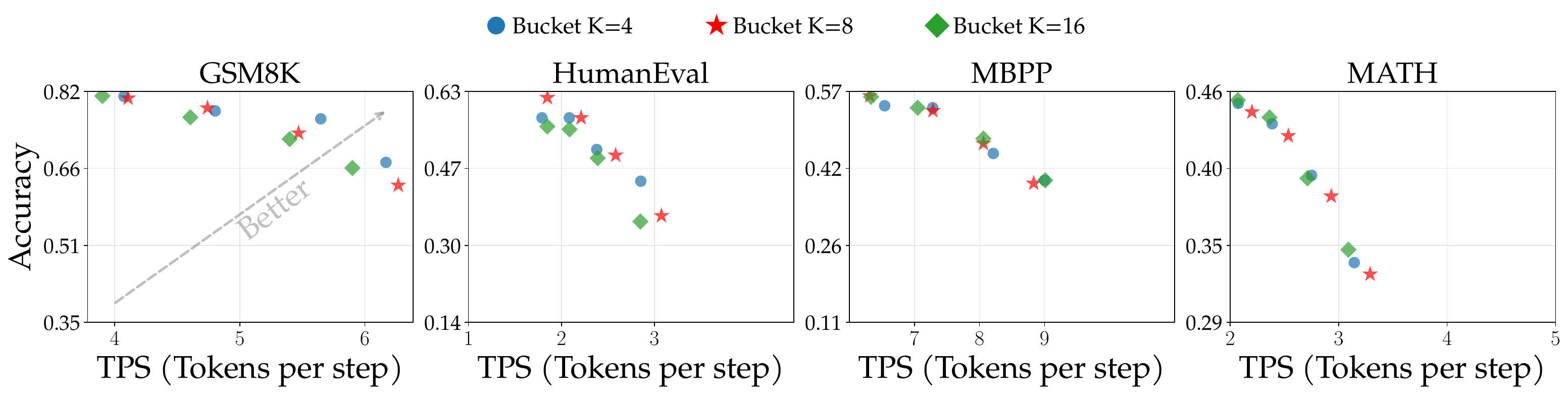}
    \caption{Dream.}
    \label{fig:ablation-bucket-dream}
  \end{subfigure}
  \caption{Effect of bucket granularity with $K \in \{4, 8, 16\}$ difficulty buckets. $K=8$ achieves the best performance, while all settings yield similar results, indicating robustness to this hyperparameter.}
  \label{fig:ablation-bucket}
\end{figure}

\section{Effect of Difficulty Metrics}
\label{appendix:difficulty-metrics}

\begin{figure}[ht]
  \centering
  \begin{subfigure}[t]{0.95\linewidth}
    \centering
    \includegraphics[width=\linewidth]{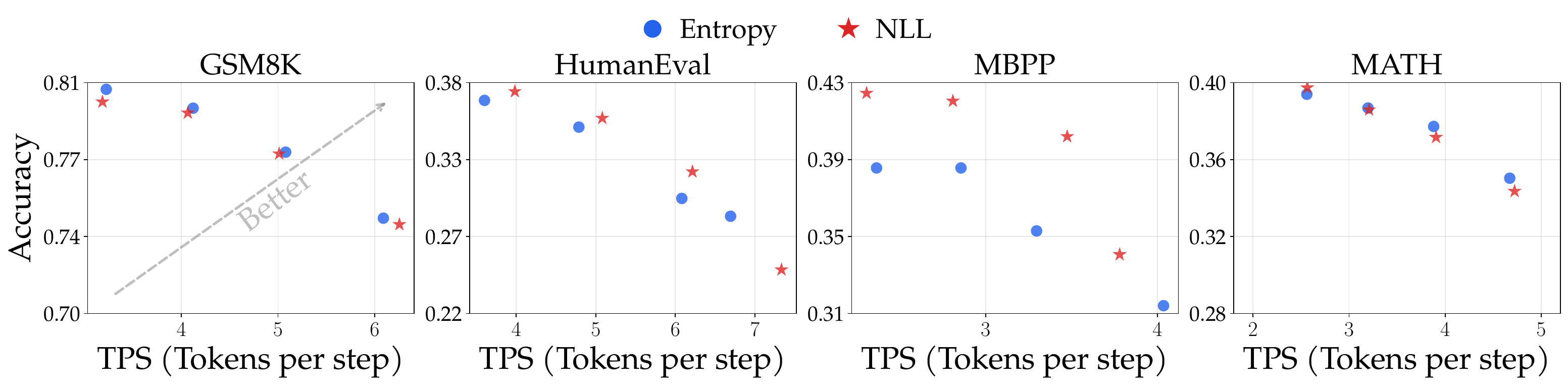}
    \caption{LLaDA.}
    \label{fig:ablation-metric-llada}
  \end{subfigure}

  \begin{subfigure}[t]{0.95\linewidth}
    \centering
    \includegraphics[width=\linewidth]{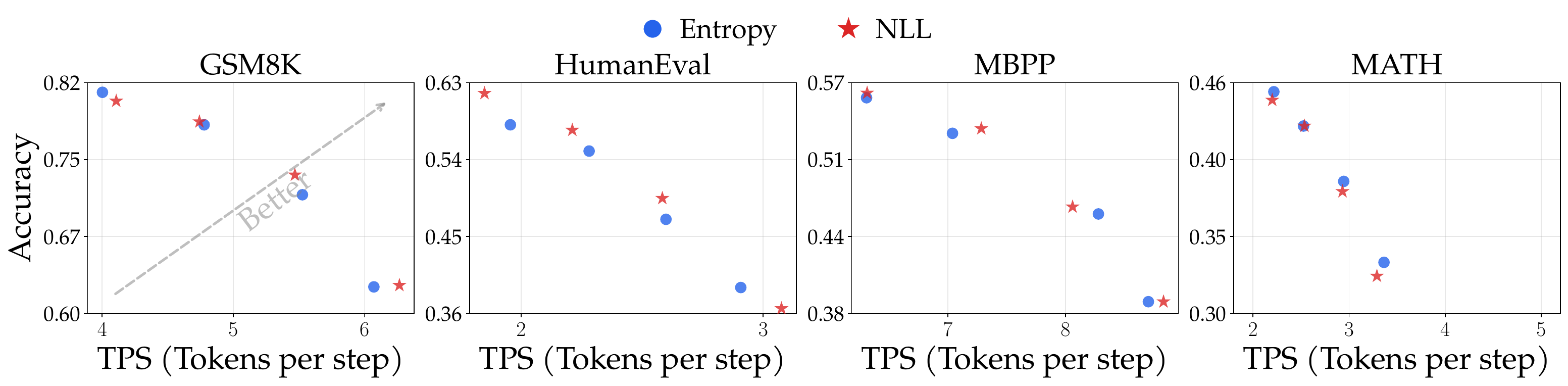}
    \caption{Dream.}
    \label{fig:ablation-metric-dream}
  \end{subfigure}
  \caption{Comparison of NLL and entropy as difficulty metrics. The two metrics produce comparable results on math benchmarks, while NLL yields a noticeably better accuracy-parallelism trade-off on coding tasks.}
  \label{fig:ablation-metric}
\end{figure}

\section{Training Cost Calculation}
\label{appendix:training-cost}
We estimate the training cost of distillation-based methods (d3LLM~\citep{qian2026d3llm} and dParallel~\citep{chen2025dparallel}) based on their reported dataset construction procedure, which we summarize below for completeness. Note that d3LLM uses the same distillation dataset as dParallel, so the curation cost is shared.

The dataset is constructed by generating teacher trajectories using a semi-autoregressive remasking strategy~\citep{qian2026d3llm}: for each prompt, the teacher produces a response trajectory, which is split into contiguous blocks. Training examples are then created by masking an ``active block'' while keeping earlier blocks as context and later blocks fully masked, simulating intermediate decoding states for certainty-forcing distillation.

For LLaDA-8B-Instruct~\citep{nielarge}, prompts are sampled from GSM8K~\citep{cobbe2021gsm8k}, PRM12K~\citep{lightman2023let}, and a subset of Numina-Math~\citep{li2024numinamath}. The teacher generates trajectories with sequence length $256$ and block length $32$, followed by filtering for correctness, yielding approximately 92K samples. For Dream-7B-Instruct~\citep{ye2025dream}, the same strategy is applied with additional code data from a subset of AceCode~\citep{zeng2025acecoderacingcoderrl} (${\sim}$10K prompts). Notably, for Dream, standard random masking is used during training instead of semi-autoregressive masking, as the latter risks degenerating toward autoregressive behavior~\citep{qian2026d3llm}.

All training tokens in these methods are self-generated without external target completions. All GPU hour estimates are based on NVIDIA A100 80GB GPUs. By contrast, \method{} uses only 1K existing prompt-response pairs with no trajectory generation, making the curation cost negligible.

\section{Related Work}

\paragraph{Diffusion Language Models.}
Diffusion models, originally developed for continuous domains~\citep{sohl2015deep,ho2020denoising,song2020score}, have been extended to discrete text via absorbing-state~\citep{austin2021structured} and uniform-state~\citep{hoogeboom2021argmax} formulations. Continuous-time extensions~\citep{campbell2022continuous}, score-based~\citep{sun2022score,meng2022concrete}, and ratio-based~\citep{lou2023discrete} objectives further unify the theory. Masked diffusion language models (MDLMs) simplify the forward process to independent token masking, with recent work clarifying equivalences and simplifying training~\citep{sahoo2024simple,shi2024simplified,ou2024your,zheng2024masked}. Alternative directions include continuous diffusion in embedding space~\citep{li2022diffusion,gong2022diffuseq,dieleman2022continuous,lin2023text}, flow matching and edit-based methods~\citep{gat2024discrete,havasi2025edit,nguyen2025oneflow}, block diffusion that interpolates between AR and diffusion decoding~\citep{arriolablock}, and hybrid AR--diffusion architectures for speculative drafting~\citep{christopher2025specdiff}, planned generation~\citep{israel2025planned}, or unified draft-and-verify passes~\citep{liu2025tidar}.
More recent works have scaled DLMs substantially. DiffuLLaMA~\citep{gong2024scaling} demonstrates that pretrained AR backbones can be adapted into DLMs through continual pretraining. LLaDA and Dream further show that DLMs can reach billion-parameter regimes and instruction-following settings, making them competitive on reasoning and coding tasks~\citep{nielarge,ye2025dream}. However, a growing body of analysis highlights that the practical benefits of parallel decoding are still limited: current DLMs exhibit strong locality bias and sensitivity to appended mask tokens~\citep{horvitz2026rethinking,piskorz2026masks}, and standard supervision can induce autoregressive-like decoding dynamics even in nominally parallel DLMs~\citep{li2026why}. These findings motivate methods that explicitly improve the decoding trajectory during training.

\paragraph{Trajectory Optimization via RL and Distillation.}
A closely related line of work attempts to improve diffusion decoding trajectories directly. Distillation-based methods such as dParallel~\citep{chen2025dparallel} and d3LLM~\citep{qian2026d3llm} train DLMs to follow more aggressive parallel decoding schedules, often by imitating or compressing sampled trajectories so that more tokens can be resolved at earlier denoising steps. Reinforcement-learning-based approaches have also been proposed for DLMs. CJ-GRPO~\citep{yang2025taming} introduces consistency-aware trajectory reinforcement learning for fewer-step decoding, while SPG~\citep{wang2026spg} develops a policy-gradient estimator tailored to DLMs with intractable likelihoods. These methods demonstrate that decoding trajectories are a critical lever for improving the accuracy--parallelism trade-off. However, they typically rely on online diffusion sampling, sampled pseudo-trajectories, or trajectory-level optimization, which introduce substantial computational overhead.

% \paragraph{Efficient Training and Supervision Signals.}
% Our method is also related to a broader class of efficient supervision strategies that seek to transfer useful structure without expensive online optimization. Knowledge distillation shows that teacher predictions can provide richer supervision than standard hard targets, while sequence-level distillation extends this idea to structured generation settings \citep{hinton2015distilling,kim2016sequence}. Curriculum learning further suggests that exposing models to training signals in an organized order can improve optimization and generalization \citep{bengio2009curriculum}. In the diffusion-language-model setting, adaptation-based methods such as DiffuLLaMA already exploit autoregressive models as a source of efficient initialization \citep{gong2024scaling}. Our approach is complementary to these efforts: instead of distilling full trajectories or using costly diffusion sampling, we extract token-level difficulty signals from an autoregressive teacher with a single teacher-forcing pass, and use them to construct trajectory-aware supervision for diffusion training.

\end{document}